\documentclass{article}

\usepackage[includeheadfoot,
            bindingoffset=0mm,
            inner   = 25mm,
            top     = 10mm,
            outer   = 25mm,
            bottom  = 10mm,
            paperwidth = 210mm,
            paperheight = 297mm,
            ]{geometry}
\usepackage[margin={2.0cm,0cm},oneside,labelfont={sf,bf},singlelinecheck=false]{caption}

\usepackage{graphicx}

\usepackage[fleqn]{amsmath}
\usepackage{accents}
\usepackage{amssymb}

\usepackage{enumitem}
\usepackage{cite}
\usepackage[perpage,multiple]{footmisc}
\usepackage{hyphenat}
\usepackage[english]{babel}
\usepackage[section]{placeins}
\usepackage{color}
\usepackage{multirow}

\usepackage{titlesec}
\titleformat{\section}[hang]{\Large\bfseries\raggedright\sffamily}{\thesection}{1em}{}
\titleformat{\subsection}[hang]{\large\bfseries\raggedright\sffamily}{\thesubsection}{1em}{}
\titleformat{\subsubsection}[hang]{\normalsize\bfseries\raggedright\sffamily}{\thesubsubsection}{1em}{}
\usepackage{abstract}

\usepackage{authblk}

\newcommand{\vectorn}[1]{\ensuremath{ \boldsymbol{#1} }}

\begin{document}

\title{\huge\bfseries\sffamily Concurrent training methods for Kolmogorov-Arnold networks: Disjoint datasets and FPGA implementation}

\author[1]{Andrew Polar}
\author[2,3]{Michael Poluektov}
\affil[1]{Independent software consultant, Duluth, GA, USA}
\affil[2]{School of Computing and Mathematical Sciences, University of Greenwich, Park Row, London SE10 9LS, UK}
\affil[3]{Corresponding author, email: m.poluektov@greenwich.ac.uk}

\date{ \large\normalfont\sffamily DRAFT: \today }

\maketitle

\setlength{\absleftindent}{2.0cm}
\setlength{\absrightindent}{2.0cm}
\setlength{\absparindent}{0em}

\begin{abstract}
The present paper introduces concurrency-driven enhancements to the training algorithm for the Kolmogorov-Arnold networks (KANs) that is based on the Newton-Kaczmarz (NK) method. Prior research shows that KANs trained using the NK-based approach outperform classical neural networks (multilayer perceptrons --- MLPs) both in terms of accuracy and training time. Up to now, the fundamental limitation of the algorithm has been the sequential computation of the updates --- each update depends on the results of the previous step, obstructing parallelisation; even though parallelisation of some parts of the algorithm, such as the evaluation of the basis functions, has already been proposed and tested. However, substantial acceleration is achievable. Three complementary concurrency-driven novel strategies are proposed in the present paper: (i) a pre-training procedure tailored to the NK updates' structure, (ii) training on disjoint subsets of data together with models' merging, and (iii) a division-free customisation of the algorithm for field-programmable gate arrays (FPGAs), which is implemented and tested directly on the device. Computational experiments are used to assess the scalability and to compare the authors' KAN implementation with popular MLP packages. All presented experimental results are fully reproducible, with the complete source codes available online.
\end{abstract}

\section{Introduction}
\label{sec:intro}

A Kolmogorov-Arnold network (KAN) is a non-linear regression model, i.e.\ a ``black box'' that converts an input vector to an output vector, alternative to a classical neural network --- multilayer perceptron (MLP). There are important differences between KANs and MLPs, discussion of which the reader can find in abundant literature on this topic. Recent popularity of KANs can be attributed to \cite{Liu2024}. However, KANs were not invented in 2024 and originate from the pioneering work of Igelnik and Parikh \cite{Igelnik2003} published in 2003. A brief historical overview of KANs' development can be found in e.g.\ \cite{Poluektov2023} and references therein.

To construct models based on KANs, one needs to make a number of choices. In particular, KANs contain basis functions (a rough analogy may be activation functions in MLPs), which can be chosen to be splines, Gaussians, piecewise-linear functions, other polynomials, etc. One then needs to choose a training method (also called `optimiser'), which fits the parameters of the model to the given dataset. Other important choices include the number of layers in the model (again, a rough analogy may be the number of layers in MLPs), the number of blocks within each layer, and the number of basis functions per underlying function of each block. 

Following \cite{Liu2024}, nowadays, it is popular to combine the spline basis functions with `Adam' or LBFGS training methods. However, the primary consequence of such choice is relatively long training time. Earlier approaches \cite{Montanelli2020,Polar2021} advocated the use of piecewise-linear basis functions, which can achieve the same accuracy within a significantly shorter training time (for models that do not require the continuity of the derivatives of the outputs by the inputs). In \cite{Polar2021}, a training algorithm based on the Kaczmarz method (i.e.\ a method for solving linear systems of equations) has been proposed; the algorithm has been improved in \cite{Poluektov2023}, where it has also been shown that the approach has a significant performance edge. However, the approach relies on a series of sequential steps, the full parallelisation of which has not been resolved up to now.

The aim of the present paper is to propose four novel strategies that are primarily concurrency-oriented and improve the training time of an already efficient training algorithm for KANs based on the Newton-Kaczmarz (NK) method. The strategies are (i) an adaptive algorithm for numerical damping, (ii) a new pre-training procedure, (iii) a concurrent training approach using disjoint subsets of the dataset together with the subsequent merging of the models, (iv) an integer-based customisation of KANs together with the training algorithm suitable for field-programmable gate arrays (FPGAs). The paper demonstrates the first implementation of KANs' training on FPGAs --- the initial demo\footnote{https://github.com/andrewpolar/fpga} in SystemVerilog code was completed and shared via GitHub in January 2026. The overall novelty of the paper is twofold --- it consists of (a) the parallelisation of the NK-based training algorithm and (b) the adaptation and implementation of the algorithm on FPGAs.

Strategies (ii)-(iv) focus specifically on parallelisation. Strategies (i)-(iii) are general and are valid for any choice of the basis functions. The implementation on FPGAs however relies on piecewise-linear basis functions. Although basis functions with continuous derivatives are essential for some applications, such as physics-informed modelling, they are not universally required. Moreover, the basis used during the training does not need to coincide with the basis used in the final model. In particular, a model may be trained efficiently using a piecewise-linear representation and subsequently upgraded to splines or other bases with only minor post-training, thereby resulting in a spline-based model under substantially reduced training time\footnote{This has been proposed in the previous publication by the authors \cite{Poluektov2023}, implemented in the C++ code, and computationally tested. The example code is available at https://github.com/andrewpolar/kanpro}. Thus, selecting piecewise-linear functions does not represent a move towards a more primitive model; instead, it enables faster training while retaining full descriptive capabilities.

A notable property of KANs, independently observed by multiple authors \cite{Zeydan2024_FKANs,Sasse2024_EvaluatingF_KAN_NonIID,Ma2025_EnhancingFL_KAN,Lee2025_BenchmarkFL_KAN_MedicalImaging} and exploited in the present work, is the possibility of merging the models trained on disjoint datasets in a straightforward manner. In contrast, it is known that conventional neural networks generally lack an equally simple merging mechanism. Up to now, this property has been primarily investigated in the context of federated learning. The present work instead examines its use for concurrent training on disjoint data subsets.

The interest in FPGA implementation stems from the ability of integrated circuits (ICs) to support efficient massive parallelisation. Unlike CPU and GPU architectures, chains of parallel and sequential operations on FPGAs do not require explicit synchronisation after each parallel stage. Research on using FPGAs for KAN training is still in its early stages. To the best knowledge of the authors, as of now, only two related works exist \cite{Hoang2025KANELE,Hoang2026}. The first paper considers only inference, with the model training performed on a different device. The second publication reports successful training of KANs directly on FPGAs.

The FPGA-related contribution of the present paper has some similarities to \cite{Hoang2026} --- both employ training that is conceptually related to the method originally proposed for KANs in preprints from 2020 and 2023, later published as \cite{Polar2021,Poluektov2023}. However, the present work differs in its customisation of the model together with the training algorithm specifically for FPGA hardware. A complete description of the implementation is given below, together with the links to the accompanying register-transfer level (RTL) source code, which is released as a reusable template. In contrast, the corresponding RTL source code of \cite{Hoang2026} has not been made publicly available, making independent reproduction substantially more difficult. 

\section{The core concept of the approach}
\label{sec:NK}

A Kaczmarz-based training method for KANs with piecewise-linear basis functions was introduced by the authors in 2020 in a preprint later published as \cite{Polar2021}; the approach was generalised to the NK-based training method for KANs with arbitrary basis functions in 2023 in a preprint later published as \cite{Poluektov2023}. The reader is referred to the above publications for details; however, for completeness of the paper, a very brief summary is provided below (covering only piecewise-linear basis functions for simplicity).

\textbf{Origins of the approach.} The NK method solves systems of non-linear algebraic equations \cite{Meyn1983} that can be underdetermined or overdetermined. When applied to a linear system (also underdetermined or overdetermined) it becomes the Kaczmarz method or the projection descend method \cite{Kaczmarz1937} --- each equation corresponds to a plane in a hyperspace, the solution corresponds to an intersection of these planes, and the method iteratively projects the initial guess point from one plane onto another until it approaches the intersection point. In the case of non-linear systems, before each projection step, an additional linearisation step is performed (i.e.\ the Newton's step). 

KAN is a regression model --- a set of multivariate functions converting an input vector into an output vector. These functions have trainable parameters. Thus, having some dataset, the training problem for KAN can be formulated as a system of non-linear equations, where the non-linear functions (i.e.\ KAN) take the training dataset inputs and produce the training dataset outputs. Therefore, the core idea of the training method is to solve this non-linear system of equations with respect to the parameters of KAN using the NK method.

\textbf{Concept.} The basic building block of KAN is an operator that converts $m$-dimensional vector $\vectorn{y}$ into scalar $z$. It can be called the discrete Urysohn operator \cite{Krylov1979} or a generalised additive model \cite{Hastie1990}:
\begin{equation}
z = \sum_{j=1}^{m} g_j \left( y_j \right) ,
\label{eq:DUO}
\end{equation}
where $g_j$ are some functions with input range $y_j \in \left[y_\mathrm{min}, y_\mathrm{max} \right]$. These vector-to-scalar mappings are referred to just as `blocks' below. One layer of KAN consists of a set of such blocks, forming $n$-dimensional vector $\vectorn{z}$, such that
\begin{equation}
z_i = \sum_{j=1}^{m} g_{i,j} \left( y_j \right) , \qquad i = 1,2,\ldots,n.
\label{eq:layer}
\end{equation}
Thus, one layer converts vector $\vectorn{y}$ into vector $\vectorn{z}$. KAN is a chain of such layers, where the outputs of the previous layer are the inputs for the next layer. 

Various representations for functions $g_{i,j}$ can be used. The simplest option is to take them to be piecewise-linear. For uniformly-spaced nodes (points where the functions change the slope) with inter-nodal distance $\Delta y$, the interpolation is trivially computed via
\begin{align}
&s = 1 + \left( y_j - y_\mathrm{min} \right) / \Delta y , 
\qquad k = \lfloor s \rfloor , 
\qquad f = s - k , \label{R} \\
&g_{i,j} \left( y_j \right) = \left( 1 - f \right) G_{i,j,k} + f \, G_{i,j,k+1} , \label{eq:g-interp} 
\end{align}
where $k$ is the index of the left node, $f \in [0,1)$ is the relative offset from that node, and model parameters $G_{i,j,k}$ are also the values of functions $g_{i,j}$ at the nodes.

Training of KAN consists in updating parameters layer-by-layer for each record of the training dataset. To update one layer, a `desired' (i.e.\ `target') vector $\vectorn{z}^*$ is required. For each function $g_{i,j}$, only two nodes are updated:
\begin{align}
&G_{i,j,k}^\mathrm{new} = G_{i,j,k}^\mathrm{old} + \mu \left( z_i^* - z_i \right) \left( 1 - f \right) , \label{eq:upd1} \\
&G_{i,j,k+1}^\mathrm{new} = G_{i,j,k+1}^\mathrm{old} + \mu \left( z_i^* - z_i \right) f , \label{eq:upd2} 
\end{align}
where $k$ and $f$ depend on index $j$ according to equation \eqref{R}, multiplier $\mu$ is the scaling factor, which depends on offsets in all layers in the original form of the method.

For the outer layer of KAN, the target vector is given in the training dataset. For some given layer, if target vector $\vectorn{z}^*$ is known, then target vector $\vectorn{y}^*$ for the preceding layer is calculated as
\begin{equation}
\vectorn{y}^* = \vectorn{y} + \vectorn{J}^\mathrm{T} \left( \vectorn{z}^* - \vectorn{z} \right) ,
\label{product}
\end{equation}
where $\vectorn{J}$ is the Jacobian matrix with elements
\begin{equation}
J_{i,j} = \left( G_{i,j,k+1} - G_{i,j,k} \right) / \Delta y ,
\end{equation}
with $k$ being dependent on index $j$ according to equation \eqref{R}.

\section{Proposed improvements}
\label{sec:ID}

\subsection{Adaptive numerical damping via dynamic rescaling}

The domains of all functions in KAN, i.e.\ $y_\mathrm{min}$ and $y_\mathrm{max}$ of each layer, must be initialised before the training. It can be proven that (see appendix \ref{sec:rescaling})
\begin{itemize}
\item $y_\mathrm{min}$ and $y_\mathrm{max}$ can be arbitrary;
\item there exist optimal values of $y_\mathrm{min}$ and $y_\mathrm{max}$ that provide the fastest training;
\item changing $y_\mathrm{min}$ and $y_\mathrm{max}$ is equivalent to modifying scaling factors $\mu$ differently in different layers.
\end{itemize}
Using these three facts, one can establish two strategies for acceleration of the training.

The first strategy is to fine-tune scaling factors $\mu$ manually for each layer, such that the fastest training is achieved. 

The second strategy is to introduce a linear mapping on top of each block, replacing equation \eqref{eq:layer} by
\begin{equation}
z_i = \alpha_i \sum_{j=1}^{m} g_{i,j} \left( y_j \right) + \beta_i, \qquad 
i = 1,2,\ldots,n,
\label{eq:layer-sc}
\end{equation}
where $\alpha_i$ and $\beta_i$ are some constants that are allowed to change dynamically and differently for each component of vector $\vectorn{z}$. Initially, $\alpha_i = 1$ and $\beta_i = 0$ are set. Furthermore, the initial expectation is that components $z_i$ vary between $z_\mathrm{min}$ and $z_\mathrm{max}$, which is the predefined range of the input of the next layer. As the training proceeds, the model parameters evolve, and when $z_i$ becomes smaller than $z_\mathrm{min}$ or larger than $z_\mathrm{max}$, values $\alpha_i$ and $\beta_i$ are updated such that $z_i$ falls back between $z_\mathrm{min}$ and $z_\mathrm{max}$. After the training is finished, the model parameters are rescaled, i.e.\ the values of $\alpha_i$ and $\beta_i$ are pulled into parameters $G_{i,j,k}$, such that $\alpha_i$ and $\beta_i$ are no longer needed and KAN's layers revert back to formula \eqref{eq:layer}.

Thus, the linear mapping is temporary and is introduced only for the training stage. It is equivalent to adapting the domains of the functions dynamically, just values of $\alpha_i$ and $\beta_i$ are changing instead of $z_\mathrm{min}$ and $z_\mathrm{max}$. From the proof of appendix \ref{sec:rescaling}, it follows that such procedure is also equivalent to adapting scaling factors $\mu$ dynamically, while keeping the functions' domains fixed.

\subsection{Pre-training}
\label{sec:pre}

The classical vector-input-scalar-output two-layer KAN is the following model:
\begin{equation}
z = \sum_{j=1}^{m} g_j \left( y_j \right) , \qquad
y_j = \sum_{l=1}^{q} h_{j,l}\left(x_l\right) ,
\label{eq:addends}
\end{equation}
where $x_l$ are the model inputs (components of input vector $\vectorn{x}$) and $z$ is the output scalar. The left and the right equations correspond to the outer and the inner layers, respectively. The parameters are initialised at random. The novel pre-training concept for two-layer models consists in creating models comprising of smaller number of addends of the outer layer:
\begin{equation*}
z^{(1)} = \sum_{j=1}^{v} g_j \left( y_j \right), \qquad
z^{(2)} = \sum_{j=v+1}^{2v} g_j \left( y_j \right), \qquad
\ldots \qquad
z^{(m/v)} = \sum_{j=m-v+1}^{m} g_j \left( y_j \right). \qquad
\end{equation*}
These models are trained concurrently, each approximating target $z^*$. After the training, these models are assembled into original model \eqref{eq:addends} with the appropriate scaling (by $m/v$), simply by collating all the addends together. Such approach fully employs multi-threading, as computational threads (one model $z^{(i)}$ per thread) can work independently. 

In the case of a three-layer model, the main idea of the pre-training consists in training a classical (two-layer) model first, then disregarding the outer layer and taking the intermediate variable values as the new inputs, using which another classical (two-layer) model is trained. This gives the initial approximation for three layers, which are then trained in a standard way. Each classical (two-layer) model can then be trained either from the initial random state or with the described-above pre-training algorithm for the two layers.

Analogously, pre-training for an arbitrary multi-layer version of the model can be done via a loop, where at each step the following is performed: two layers are trained, the outer layer is disregarded, the inner layer is used to create new input vectors. Each two-layer model can then be pre-trained as above.

\subsection{Training on disjoints and merging}

The described approach iterates record-by-record through the training dataset. The novel parallelisation strategy consists in splitting the training dataset into subsets, which are used to train several copies of the model in parallel. Each subset should be sufficiently representative of the modelled system, i.e.\ it is implied that the algorithm is used for large data. The proposed strategy represents the following sequence:
\begin{enumerate}[label=(\alph*)]
\item create $p$ copies of the model (KAN);\label{it:begin}
\item form $p$ equal-size subsets (batches) using the records from the training dataset;
\item train each copy of the model on the corresponding batch (one pass through the data) \emph{in parallel};
\item merge all models into a single model simply by computing the mean value of every parameter $G_{i,j,k}$;\label{iter:merg}
\item go to step \ref{it:begin}.
\end{enumerate}
In this procedure, one should choose the number of batches and the batch size --- these are the only two hyperparameters that govern the training on disjoints (in contrast to federated learning, these parameters may be chosen arbitrarily). The above steps are iteratively repeated until the selected convergence criteria is reached. 

Using this approach, processing of the records from the training dataset should be parallelisable almost ideally, with minor losses for model coping/merging. However, there will be some loss of accuracy due to the averaging, and one may expect increasing error with the number of batches. Therefore, there should be an optimal number of batches. Thus, the computational experiments serve two \emph{completely separate} purposes: the first is to establish the scaling of the implementation of the algorithm (i.e.\ how much time is lost for coping/merging), the second is to find how the accuracy drops with the number of batches.

It should be noted that the novelty here is not in the idea of model merging via parameter averaging (step \ref{iter:merg} of the algorithm), which has been proposed earlier for KANs in \cite{Zeydan2024_FKANs,Sasse2024_EvaluatingF_KAN_NonIID,Ma2025_EnhancingFL_KAN,Lee2025_BenchmarkFL_KAN_MedicalImaging} to perform federated learning, but in applying this technique to accelerate the training, given that the dataset is located entirely on a single storage space.

\subsection{All-integer division-free KANs for FPGAs}
\label{FPGA-1}

\subsubsection{General concept of training on FPGAs}

Taking the two-layer KAN as an illustrative example (chosen for simplicity), according to the algorithm outlined in section \ref{sec:NK}, the computational flow for processing one record can be schematically illustrated as follows:
\begin{equation*}
\vectorn{x} \; \rightarrow \; \vectorn{y} \; \rightarrow \; \vectorn{z} \; \rightarrow \; \left( \vectorn{z}^* - \vectorn{z} \right) \;\rightarrow \; \left( \vectorn{y}^* - \vectorn{y} \right) \; \rightarrow \; \text{updated } g_{i,j} \text{ and } h_{j,l} ,
\end{equation*}
where superscript `$*$' denotes the target output of a layer, $\vectorn{x}$ is the input vector, $\vectorn{y}$ is the intermediate (hidden) vector, $\vectorn{z}$ is the output vector, $g_{i,j}$ and $h_{j,l}$ are the functions of the outer and the inner layer, respectively. 

The operations within each `arrow' are independent. The entire concept of parallelisation for FPGAs is to perform operations of each `arrow' \emph{concurrently}. The number of clock cycles required for one such `arrow' depends on the hardware and the art of programming --- it may be as low as one or two cycles, independently of the model size. More detailed workflow is given below.

The specific implementation demonstrated here uses the piecewise-linear basis functions as described in section \ref{sec:NK}. It implies that the output variables are not continuously differentiable. However, a trained piecewise-linear model can then be exported from an FPGA board, and the basis functions can be replaced by e.g.\ splines, as discussed in the Introduction.

\subsubsection{All-integer model with bit shifting and masking instead of divisions}

FPGAs use fixed-point representations of values, which closely resemble integers; therefore, the easiest option is to use exclusively integer values for all variables. Because inputs/outputs/parameters of KANs can be rescaled arbitrarily, as well as the intermediate functions' domains, as outlined in appendix \ref{sec:rescaling}, KANs are naturally suited to integer-based operations. This approach requires ensuring that overflow and accumulated loss of significant bits are avoided. 

Switching to an all-integer model also allows eliminating division operations. The most frequent division operation is used in equation \eqref{R} for determining the index of the linear segment. Due to arbitrary rescaling of $y_\mathrm{min}$ and $y_\mathrm{max}$, the segment lengths are chosen as $\Delta y = 2^d$, where $d$ is some integer. This allows implementing the division by $\Delta y$ as a binary shift by $d$ bits. Calculation of offset $f$ in equation \eqref{R} can then be implemented via application of a bit mask. The same logic applies to any scaling; for example, division by some integer $n$ can be implemented approximately as a multiplication (by $\operatorname{round}(2^d/n)$) followed by a binary shift (by $d$ bits). Multipliers $\mu$ in equations \eqref{eq:upd1} and \eqref{eq:upd2} are stored as power-of-two reciprocals, allowing the implementation of multiplication via binary shifts.

When outputs $z_i$ of a layer fall outside of the input range of the subsequent layer, these outputs are truncated (i.e.\ if $z_i > z_\mathrm{max}$, then assign $z_i = z_\mathrm{max} - \epsilon$; if $z_i < z_\mathrm{min}$, then assign $z_i = z_\mathrm{min} + \epsilon$), where $\epsilon$ is the smallest offset. Frequent violation of the ranges should of course be avoided, as it may degrade the model accuracy. As outlined above, it can be controlled by adjusting the numerical damping parameters for individual layers.

\subsubsection{Scheme of parallel implementation}

The current FPGA implementation is only a proof-of-concept; therefore, specific choices are hardcoded. The easiest way to understand the concept is by looking at the code that is available online\footnote{https://bitbucket.org/kolmogorov-arnold/det-33-c-fpga/src/master/}. The repository contains two files: a SystemVerilog code and a C++ code that uses all computational operations (e.g.\ integer arithmetic) and \emph{pseudo concurrency} as on FPGAs. The latter implies that independent operations, which are naturally executed on FPGAs in parallel, are marked by comments but are executed sequentially on a CPU. The individual operations and the final result are identical --- all intermediate print-out values between the C++ demo and the actual device match at the bit level. 


The implementation has been written for Digilent Nexys A7-100T, which is a very basic device with limited resources. For this device, the processing of one training record has been coded in exactly $14$ clock cycles (constructing a two-layer network):
\begin{itemize}
\item cycles $1$--$4$ (forward compute): for each of two layers, all functions $g_{i,j}$ from equation \eqref{eq:g-interp} are calculated \textit{in parallel} within the one cycle, and the sum from equation \eqref{eq:layer} is evaluated within the next cycle;
\item cycles $5$--$6$ (evaluation of the residuals): for the outer layer, residuals $z_i^* - z_i$ are calculated within one cycle; for the inner layer, the residuals are calcualted via equation \eqref{product} within the next cycle;
\item cycles $7$--$10$ (evaluation of the updates): for each of two layers, all $\mu \left( z_i^* - z_i \right) f$ are calculated \textit{in parallel} within the one cycle, and all $\mu \left( z_i^* - z_i \right) \left( 1 - f \right)$ are calculated \textit{in parallel} within the next cycle;
\item cycles $11$--$14$ (applying the updates): for each of two layers, all $G_{i,j,k+1}$ from equation \eqref{eq:upd2} are updated \textit{in parallel} within the one cycle, and all $G_{i,j,k}$ from equation \eqref{eq:upd1} are updated \textit{in parallel} within the next cycle.
\end{itemize}
From the structure of the scheme, it is clearly seen that the sizes of the layers are not affecting the number of cycles (i.e.\ the calculation time) --- any layer sizes (up to the hardware limit) can be used and will be processed within the same number of cycles. Furthermore, the number of cycles can be potentially reduced by combining some of the cycles (for which more processing power within one parallel block will be required). Obviously, the implementation can be straightforwardly modified to models with more than two layers. 


\section{Numerical experiments}
\label{sec:BandT}

The aim of this section is to demonstrate the proposed improvements of the training procedure using simple intuitively-understandable examples. The examples are specifically based on synthetic data as it allows controlling precisely the sources of errors and the uncertainty. In particular, the data is deterministic, algebraic, and continuous; therefore, theoretically, it is possible to build $100\%$ accurate models. Furthermore, random generation of the data at each programme run prohibits tailoring of the models to specific patterns found in datasets that have fixed recorded observations. Such data allows drawing precise conclusions regarding the performance of the parallelisation techniques proposed in the present paper. The serial version of the algorithm has been extensively tested on real-world data, see e.g.\ \cite{Polar2021,Poluektov2023}; improvement/extension of the original algorithm via parallelisation does not change the validity of the algorithm and its relevance to real-world ML scenarios.

The first series of examples is the prediction of determinants of random $N \times N$ matrices. The inputs are $N^2$ elements of a matrix; the output is the determinant of the matrix. These examples provide the synthetic data that is simultaneously complex to model and conceptually simple to explain (compared to e.g.\ multivariate formulas). Three cases are considered:
\begin{itemize}
\item \emph{Det5}: $5 \times 5$ matrices, $10\mathrm{M}$ training records, $2\mathrm{M}$ validation records;
\item \emph{Det4}: $4 \times 4$ matrices, $100\mathrm{K}$ training records, $20\mathrm{K}$ validation records;
\item \emph{Det3}: $3 \times 3$ matrices, $50\mathrm{K}$ training records; online-type training where each record is unseen.
\end{itemize}
The fourth example is the prediction of the areas of triangular faces of randomly-generated tetrahedra ($4$ faces). The inputs are $12$ coordinates of the vertices of the tetrahedra in 3D ($4$ vertices, $3$ coordinates for each); the outputs are $4$ areas:
\begin{itemize}
\item \emph{Tetra}: $4$ areas of tetrahedra faces, $200\mathrm{K}$ training records, $20\mathrm{K}$ validation records.
\end{itemize}
This example tests the performance of the training procedure on a challenging dataset with multiple correlated outputs. 

The predictive performance is assessed using the Pearson correlation coefficient (in percent) between the real and the modelled output (since the output distributions are strongly concentrated around their mean, a model producing near-mean random outputs can result in small root mean square errors and provide false indication of the accuracy). The values in the tables are the averages and the standard deviations of three programme runs. When the pre-training is used, one pass over the entire training dataset is made to pre-train the models according to the algorithm of section \ref{sec:pre}. 

The authors share the specific code that has been used for the computational tests\footnote{https://bitbucket.org/kolmogorov-arnold/dfdisjointsnew/src/master/} as well as the newer more user-friendly version of the code (see link at the end of Conclusions). It is a portable Windows/Linux C++ code, fewer than $550$ lines long, with only built-in types used, and without any third-party libraries.

\subsection{Scalability tests on HPC cluster}
\label{sec:Cluster}

To benchmark the scalability of the code (training on disjoint batches across multiple threads), the tests were run on an HPC cluster. All jobs were run on a single node equipped with \emph{two} Intel(R) Xeon(R) Platinum 8358 CPUs ($32$ cores per CPU) and $256\mathrm{GB}$ RAM (in reality, each job occupied only around $3\mathrm{GB}$ RAM).

The considered example was \emph{Det5}, and pre-training was switched off. Two-layer KANs were constructed, with $200$ blocks in the first layer and $4$ points per function, $1$ block in the second layer and $16$ points per function (the total of $23{,}200$ parameters). In the single-threaded setting, without pre-training, $2$ passes through the training dataset (epochs) are required to reach the Pearson correlation of around $95\%$, which corresponds to the processing of $20\mathrm{M}$ records --- this value was used as the initial (or required) workload.

Following the conventions of scalability benchmarking, both the \emph{weak scaling} and the \emph{strong scaling} are tested. In the weak scaling tests, the number of threads is kept proportional to the total computational work. In the strong scaling test, the total computational work is fixed, and it is equally split between the threads. For weak scaling, the efficiency is measured as $E = t(1)/t(n)$, where $t(1)$ is the time required to complete the work on $1$ thread and $t(n)$ is the time required to complete the work that is $n$ times larger on $n$ threads (in the case of ideal/linear scaling, $E=1$). For strong scaling, the speedup is measured as $S = t(1)/t(n)$, where $t(1)$ is the time required to complete the work on $1$ thread and $t(n)$ is the time required to complete the same work on $n$ threads.

Here, the total computational work is the total number of processed records of the training dataset, denoted as $N$. As highlighted above, when training is performed on disjoints, there are two configurational options: the number of batches, denoted as $T$, and the batch size, denoted as $Q$. The number of batches corresponds to the number of threads, as the batches are processed in parallel. The batch size, however, is a free configurational parameter, the influence of which is tested below. When both $N$ and $T$ are fixed, $Q$ determines the total number of model synchronisations (merges), denoted as $M = N/(TQ)$.

\subsubsection{Weak scaling}

The results are shown in table \ref{tab:HPC-w}, where it is seen that the scaling is very close to linear up until $24$ threads. The efficiency is above $97\%$ when up to $16$ threads are used and only drops below $90\%$ when $32$ threads are used. Such noticeable decline of the efficiency for a large number of threads is due to a cost of model synchronisations for large model sizes of the considered example. 

The batch size was fixed at $100\mathrm{K}$, which means that $M=200$ model synchronisations were done in each calculation. The resulting accuracy of the model (obtained on the validation dataset) drops from $95\%$ for $1$ thread to $90\%$ for $12$ threads and even further for larger number of threads. The accuracy is lost due to averaging of the models. Such noticeable accuracy loss is due to a relatively large batch size, and it can be improved by the decrease of the batch size --- a separate study shown below.

\begin{table}
\begin{center}
\caption{The weak scaling test for the authors' C++ KANs' implementation.}
\label{tab:HPC-w}
\begin{tabular}{| l | c c c c c |}
\hline
threads & $1$ & $2$ & $4$ & $8$ & $12$ \\
\hline
time (s) & $406.1 \pm 1.9$ & $409.6 \pm 1.4$ & $412.6 \pm 0.4$ & $417.4 \pm 2.6$ & $414.4 \pm 0.9$ \\
accuracy (\%) & $0.948 \pm 0.001$ & $0.946 \pm 0.001$ & $0.940 \pm 0.001$ & $0.923 \pm 0.003$ & $0.903 \pm 0.012$ \\
efficiency & $1.00$ & $0.99$ & $0.98$ & $0.97$ & $0.98$ \\
\hline
threads & $16$ & $24$ & $32$ & $48$ & $64$ \\
\hline
time (s) & $419.3 \pm 2.4$ & $439.1 \pm 10.1$ & $488.1 \pm 13.4$ & $515.1 \pm 18.2$ & $519.0 \pm 5.7$ \\
accuracy (\%) & $0.880 \pm 0.018$ & $0.853 \pm 0.009$ & $0.761 \pm 0.026$ & $0.599 \pm 0.043$ & $0.429 \pm 0.033$ \\
efficiency & $0.97$ & $0.92$ & $0.83$ & $0.79$ & $0.78$ \\
\hline
\end{tabular}
\end{center}
\end{table}

\subsubsection{Strong scaling}

The results are shown in table \ref{tab:HPC-s}, where it is seen that the scaling is very close to linear up until $18$ threads, for which the speedup of $17$ times is observed. The slight drop in the speedup for a large number of threads is again due to a cost of model synchronisations.

The number of model synchronisations was fixed at $M=200$, which means that the batch size varied, such that the same total number of processed records were done in each calculation (e.g.\ for $10$ threads and the batch size of $10\mathrm{K}$, the number of processed records between model synchronisations is $10 \times 10\mathrm{K} = 100\mathrm{K}$; with $200$ synchronisations, it results in $200 \times 100\mathrm{K} = 20\mathrm{M}$ total processed records).

The resulting accuracy of the model (obtained on the validation dataset) drops from $95\%$ for $1$ thread to $80\%$ for $10$ threads and even further for larger number of threads. Again, the accuracy is lost due to averaging of the models, and it can be compensated by the further decrease of the batch size (and consequently increase of the number of model synchronisations).

\begin{table}
\begin{center}
\caption{The strong scaling test for the authors' C++ KANs' implementation.}
\label{tab:HPC-s}
\begin{tabular}{| l | c c c c c |}
\hline
threads & $1$ & $2$ & $3$ & $4$ & $5$ \\
\hline
time (s) & $406.1 \pm 1.9$ & $204.0 \pm 0.9$ & $137.3 \pm 0.2$ & $102.7 \pm 0.4$ & $82.6 \pm 0.7$ \\
accuracy (\%) & $0.948 \pm 0.001$ & $0.940 \pm 0.002$ & $0.925 \pm 0.007$ & $0.903 \pm 0.010$ & $0.904 \pm 0.001$ \\
speedup & $1.00$ & $1.99$ & $2.96$ & $3.95$ & $4.92$ \\
\hline
threads & $6$ & $7$ & $8$ & $10$ & $12$ \\
\hline
time (s) & $69.4 \pm 0.3$ & $59.4 \pm 0.3$ & $51.8 \pm 0.4$ & $41.2 \pm 0.1$ & $35.0 \pm 0.1$ \\
accuracy (\%) & $0.868 \pm 0.005$ & $0.864 \pm 0.004$ & $0.831 \pm 0.007$ & $0.796 \pm 0.021$ & $0.759 \pm 0.023$ \\
speedup & $5.86$ & $6.84$ & $7.84$ & $9.87$ & $11.62$ \\
\hline
threads & $14$ & $16$ & $18$ & $20$ & \\
\hline
time (s) & $29.7 \pm 0.3$ & $26.6 \pm 0.3$ & $23.8 \pm 0.1$ & $22.8 \pm 2.1$ & \\
accuracy (\%) & $0.723 \pm 0.030$ & $0.701 \pm 0.002$ & $0.659 \pm 0.007$ & $0.652 \pm 0.008$ & \\
speedup & $13.66$ & $15.28$ & $17.05$ & $17.85$ & \\
\hline
\end{tabular}
\end{center}
\end{table}

\subsubsection{Influence of batch size}

The results are shown in tables \ref{tab:HPC-b1} and \ref{tab:HPC-b2}, where it is seen the decrease of the batch size leads to the increase of the model accuracy. For the simulations in tables \ref{tab:HPC-b1} and \ref{tab:HPC-b2}, the number of threads was fixed at $10$ and $20$, respectively. The total number of processed records was also fixed, leading to the increase of the number of model synchronisations with the decrease of the batch size. For $10$ threads, the model accuracy of $92\%$ was achieved with the batch size of $625$ records (the total of $3200$ model synchronisations), still providing the speedup close to $9$. For $20$ threads, the model accuracy of $90\%$ was achieved with the batch size of $200$ records (the total of $5000$ model synchronisations), still providing the speedup of $12$. The drop of speedup is of course due to the increase of the number of costly model synchronisations.

Since the model accuracy drops due to the averaging, the standard speedup calculation can be misleading --- it compares sequential and parallel processing of the same computational work. A smaller total computational work is needed in the sequential case to achieve the same accuracy as in the parallel case. In particular, for the considered example, processing of $8\mathrm{M}$ training records sequentially leads to the model with the accuracy of $0.923 \pm 0.002$, which is achieved in $162.3 \pm 0.7 \, \mathrm{s}$ (result is not shown in the tables). Given that for $10$ threads, the accuracy of $92\%$ was achieved in $46.8 \, \mathrm{s}$ (processing of $20\mathrm{M}$ training records), the effective speedup is $3.47$ when the models are trained to the same accuracy.

It should be remarked that such small batch size is not always necessary. If pre-training is switched on (it was switched off for the HPC tests above), then the starting model for the main training is not random, which gives faster convergence and also allows using larger batch sizes. This \emph{Det5} benchmark demo can also be executed on a laptop --- it is a portable Windows/Linux C++ code (see link above). The default configuration is set to use $3$ threads with pre-training switched on and with the batch size of $1.3\mathrm{M}$. The execution time varies depending on hardware from $70\,\mathrm{s}$ to $120\,\mathrm{s}$ to reach the accuracy of $0.94$--$0.95$ on the validation dataset. 

\begin{table}
\begin{center}
\caption{The variation of the batch size for $10$ threads.}
\label{tab:HPC-b1}
\begin{tabular}{| l | c c c c |}
\hline
batch size & $10\mathrm{K}$ & $5\mathrm{K}$ & $2.5\mathrm{K}$ & $1.25\mathrm{K}$ \\
\hline
time (s) & $41.2 \pm 0.1$ & $41.9 \pm 0.6$ & $43.0 \pm 0.1$ & $43.9 \pm 0.4$ \\
accuracy (\%) & $0.796 \pm 0.021$ & $0.854 \pm 0.019$ & $0.900 \pm 0.007$ & $0.912 \pm 0.002$ \\
speedup & $9.87$ & $9.69$ & $9.45$ & $9.26$ \\
\hline
batch size & $625$ & $400$ & $200$ & \\
\hline
time (s) & $46.8 \pm 1.2$ & $50.0 \pm 0.2$ & $59.0 \pm 0.5$ & \\
accuracy (\%) & $0.920 \pm 0.001$ & $0.921 \pm 0.002$ & $0.927 \pm 0.001$ & \\
speedup & $8.68$ & $8.11$ & $6.88$ & \\
\hline
\end{tabular}
\end{center}
\end{table}

\begin{table}
\begin{center}
\caption{The variation of the batch size for $20$ threads.}
\label{tab:HPC-b2}
\begin{tabular}{| l | c c c |}
\hline
batch size & $5\mathrm{K}$ & $2.5\mathrm{K}$ & $1.25\mathrm{K}$ \\
\hline
time (s) & $22.8 \pm 2.1$ & $21.9 \pm 0.1$ & $23.0 \pm 0.4$ \\
accuracy (\%) & $0.652 \pm 0.008$ & $0.737 \pm 0.006$ & $0.808 \pm 0.011$ \\
speedup & $17.85$ & $18.55$ & $17.66$ \\
\hline
batch size & $625$ & $200$ & $100$ \\
\hline
time (s) & $25.3 \pm 0.1$ & $33.8 \pm 2.8$ & $46.5 \pm 6.1$ \\
accuracy (\%) & $0.877 \pm 0.004$ & $0.902 \pm 0.003$ & $0.908 \pm 0.000$ \\
speedup & $16.07$ & $12.01$ & $8.73$ \\
\hline
\end{tabular}
\end{center}
\end{table}

\subsection{Comparison to neural networks (MATLAB, Keras, PyTorch) and another KAN implementation (FastKAN)}

\subsubsection{Computational example Det4}

In the previous publication by the authors \cite{Poluektov2023}, it has been shown that the authors' implementation is not only faster than alternative KAN implementations but is also faster than neural networks (industry-standard implementation in MATLAB has been taken) --- contrary to the popular belief that KANs are slow. For that comparison, a fully sequential authors' code has been used. It is now useful to strengthen this result by comparing neural networks and an alternative KAN implementation to the parallel version of the authors' code.

The tests were run on a Windows laptop (CPU: Intel Core i7 12700H, 14 cores; GPU: Nvidia GeForce RTX 3050 Ti Laptop). Implementations of neural networks in MATLAB (version R2023a) and Keras (TensorFlow 2.20.0 and Keras 3.12.0) were taken for the comparison, as well as the FastKAN package\footnote{https://github.com/ziyaoli/fast-kan} \cite{Li2024FastKAN}. The considered example was \emph{Det4}. 

\textbf{Comparison plan/logic.} There are two primary characteristics of models' training process: the \textit{training time} and the \textit{accuracy} of the resulting model. There also multiple secondary characteristics, e.g.\ the total number of parameters, memory footprint during the training, etc. Here, the decision has been taken to try to train different models to the same \textit{accuracy} and to compare the \textit{training time}. In such approach, there is a challenge to quantify the non-uniformity of the training process --- rapid growth of the accuracy within a short time and subsequent slow creep of the accuracy over a long time. Therefore, three representative cases are considered --- the training is done to the accuracies of $0.95$, $0.97$, and $0.99$, which correspond to a coarse model that can be obtained relatively quickly, a good-enough model, and a fine model that requires considerable training time, respectively.

\textbf{Authors' KANs.} Two-layer KANs were constructed, with $70$ blocks in the first layer and $4$ points per function, $1$ block in the second layer and $16$ points per function (the total of $5{,}600$ parameters). Three cases were considered: single-threaded without pre-training, multi-threaded without pre-training, and multi-threaded with pre-training\footnote{In pre-training, pairs of addends (see section \ref{sec:pre}) were trained, $1$ full pass through the dataset was made for each pair, pairs were trained in parallel using multi-threading.}. In the multi-threaded setting, the batch size of $25\mathrm{K}$ was used. The selected configuration was found experimentally by variation of the numbers of blocks/points/batches. To train the models to different threshold accuracies, the number of passes through the training dataset was varied. 

The results are summarised in table \ref{tab:Cpp-our}. It can be seen that the model with the accuracy of $0.97$ can be obtained in $0.7\,\mathrm{s}$ when four threads and pre-training are used. If pre-training is not used, then almost double time is required to achieve that accuracy. It can also be seen that the model with the accuracy of $0.99$ can be obtained four times faster on four threads than on one (due to the smaller number of passes required to achieve the threshold accuracy in this particular case).

\begin{table}
\begin{center}
\caption{Performance of the authors' KAN implementation in \emph{Det4} example.}
\label{tab:Cpp-our}
\begin{tabular}{| l | c c c |}
\hline
\multicolumn{4}{|c|}{1 thread, no pre-training} \\
\hline
passes & $3$ & $6$ & $90$ \\
time (s) & $1.207 \pm 0.019$ & $2.319 \pm 0.021$ & $34.844 \pm 1.477$ \\
accuracy & $0.951 \pm 0.003$ & $0.971 \pm 0.002$ & $0.989 \pm 0.001$ \\
\hline
\multicolumn{4}{|c|}{4 threads, no pre-training} \\
\hline
passes & $6$ & $12$ & $70$ \\
time (s) & $0.849 \pm 0.026$ & $1.375 \pm 0.034$ & $8.764 \pm 0.037$ \\
accuracy & $0.950 \pm 0.004$ & $0.972 \pm 0.003$ & $0.990 \pm 0.001$ \\
\hline
\multicolumn{4}{|c|}{4 threads, with pre-training} \\
\hline
passes & $2$ & $5$ & $50$ \\
time (s) & $0.373 \pm 0.011$ & $0.708 \pm 0.010$ & $6.333 \pm 0.092$ \\
accuracy & $0.952 \pm 0.001$ & $0.971 \pm 0.003$ & $0.990 \pm 0.001$ \\
\hline
\end{tabular}
\end{center}
\end{table}

\textbf{MATLAB's neural networks.} Different configurations of MATLAB's neural networks were tested with the aim of finding the best-performing ones --- the cases of $1$, $2$, and $3$ hidden layers were considered. For each case, the number of training iterations was fixed, and the number of neurons per layer was varied until the threshold accuracies were reached\footnote{They were varied with the step sizes of $5$, $10$, and $20$ for the cases of $1$, $2$, and $3$ hidden layers, respectively.}, establishing the smallest configuration for the fixed number of hidden layers. Separately, tests showed that the best accuracy is obtained with \texttt{tanh} activation function, which was selected. MATLAB's \texttt{fitrnet} function was used to train the model. The single-precision floating-point format was used for the data (compared to the double-precision format used previously in \cite{Poluektov2023}), which noticeably accelerates the training.

The results are summarised in table \ref{tab:MatlDet4}, where various configurations are shown. The `config.'\ cell contains three numbers: the number of neurons per layer, the number of hidden layers, and the number of training iterations\footnote{For example, ``$50 \times 2, 1000$'' means that \texttt{`LayerSizes',[50 50]} and \texttt{`IterationLimit',1000} are passed to \texttt{fitrnet} function.} in that order. The `num.\ par.'\ cell gives the total number of parameters for the corresponding configuration. It can be seen that the model with the accuracy of $0.97$ requires at least $51\,\mathrm{s}$ of training time (if the best configuration is used) --- this is $72$ times more than for the authors' KAN, as seen from table \ref{tab:Cpp-our}. To train the model that has the accuracy of $0.99$, at least $309\,\mathrm{s}$ are required --- $49$ times more than for the authors' KAN.

It is interesting to run MATLAB's neural networks on a GPU, using \texttt{gpuArray} conversion for the data. Configuration ``$50 \times 2, 1000$'' was taken as an indicative; its execution time on the GPU was $17.517 \pm 0.997\,\mathrm{s}$, which is only approximately $3.5$ times faster than the same case on the CPU. Obviously, one cannot objectively compare a CPU code with a GPU code, as they run on \emph{completely different} hardware. However, the authors believe that their setup is representative of a typical modern researcher's laptop, and it is beneficial to see the CPU/GPU difference in this particular case. It is surprising to notice that even the single-threaded authors' KAN implementation executed on the CPU is $7.5$ times faster than neural networks trained on the GPU ($2.319\,\mathrm{s}$ vs.\ $17.517\,\mathrm{s}$, when obtaining the model with the accuracy of $0.97$).

A disclaimer must be placed that results of the tests are most likely affected by being run on a Windows laptop. First, there is no control over the background processes, even though the effort has been taken to eliminate them. Second, there is no control over CPU heating --- in the tests the CPU temperature was around $100\,^\circ\mathrm{C}$ and the frequency was fluctuating; the CPU definitely could not operate at its maximum frequency of $4.70\,\mathrm{GHz}$, and from occasional monitoring, it was observed that the frequency stayed mostly in $3$--$4\,\mathrm{GHz}$ range. Finally, although the authors put significant effort into finding the best-performing neural network configurations, the possibility of existence of better performing configurations cannot be excluded, i.e.\ fine-tuning the number of neurons in each layer and the number of training iterations may lead to better performance; however, it is difficult to see reasons for qualitative change of the picture --- it is highly unlikely that performance gains of $40$ times can be obtained by such fine-tuning.

\begin{table}
\begin{center}
\caption{Performance of the MATLAB's neural networks in \emph{Det4} example.}
\label{tab:MatlDet4}
\begin{tabular}{| l | c c c |}
\hline
config. & $25 \!\times\! 1,\: 1000$ & $15 \!\times\! 2,\: 1000$ & $10 \!\times\! 3,\: 1000$ \\
num.\ par. & $451$ & $511$ & $401$ \\
time (s) & $16.413 \pm 0.619$ & $22.472 \pm 0.943$ & $22.869 \pm 0.252$ \\
accuracy & $0.950 \pm 0.002$ & $0.953 \pm 0.001$ & $0.948 \pm 0.004$ \\
\hline
config. & $90 \!\times\! 1,\: 1000$ & $50 \!\times\! 2,\: 1000$ & $40 \!\times\! 3,\: 1000$ \\
num.\ par. & $1621$ & $3451$ & $4001$ \\
time (s) & $50.896 \pm 1.293$ & $60.828 \pm 0.943$ & $69.869 \pm 1.877$ \\
accuracy & $0.969 \pm 0.001$ & $0.971 \pm 0.001$ & $0.971 \pm 0.001$ \\
\hline
config. & $260 \!\times\! 1,\: 2000$ & $140 \!\times\! 2,\: 2000$ & $100 \!\times\! 3,\: 2000$ \\
num.\ par. & $4681$ & $22261$ & $22001$ \\
time (s) & $309.456 \pm 0.290$ & $390.931 \pm 9.586$ & $409.930 \pm 16.808$ \\
accuracy & $0.990 \pm 0.001$ & $0.990 \pm 0.001$ & $0.989 \pm 0.001$ \\
\hline
\end{tabular}
\end{center}
\end{table}

\textbf{Keras.} The authors' tests showed that on average neural networks implemented in MATLAB are performing better than ones implemented in Keras --- the training times are slightly smaller and the accuracy is slightly better. This motivated the thorough experimentation with the network configurations in MATLAB. Therefore, for Keras, four indicative configurations were selected only to see the qualitative picture: $3$ hidden layers were used in all cases, \texttt{relu} activation function was used, the sizes of the layers are given in the results table. Each configuration was trained for $100$ passes through the training dataset, and the accuracy on the validation dataset was printed out at different pass number. The results are summarised in table \ref{Keras-CPU}, where the last row is the final training time for all $100$ passes. It can be seen that the results are somewhat similar to the ones obtained with the MATLAB's neural networks.

\begin{table}
\begin{center}
\caption{Performance of the neural networks implemented in Keras in \emph{Det4} example.}
\label{Keras-CPU}
\begin{tabular}{| l | l | r | r | r | r | } 
\hline
\multicolumn{2}{|l|}{config.} & $[256,64,16]$ & $[128,64,32]$ & $[64,64,32]$ & $[32,32,32]$ \\ 
\multicolumn{2}{|l|}{num.\ par.} & $21857$ & $12545$ & $7361$ & $2689$ \\ 
\hline
\multicolumn{2}{|c|}{} & \multicolumn{4}{|c|}{accuracy} \\
\hline
\multirow{7}{*}{\rotatebox[origin=c]{90}{pass}} & $1$ & $0.457$ & $0.443$ & $0.325$ & $0.279$ \\
& $2$ & $0.873$ & $0.780$ & $0.674$ & $0.454$ \\
& $5$ & $0.899$ & $0.893$ & $0.865$ & $0.714$ \\
& $10$ & $0.905$ & $0.900$ & $0.889$ & $0.828$ \\
& $20$ & $0.918$ & $0.910$ & $0.899$ & $0.862$ \\
& $50$ & $0.967$ & $0.935$ & $0.922$ & $0.877$ \\
& $100$ & $0.975$ & $0.964$ & $0.954$ & $0.886$ \\
\hline
\multicolumn{2}{|l|}{time (s)} & $31.28$ & $27.59$ & $24.11$ & $22.33$ \\
\hline
\end{tabular}
\end{center}
\end{table} 

\textbf{FastKAN.} The FastKAN package provides an alternative implementation of KANs and is typically regarded as the fastest implementation by the community. Three different four-layer configurations with different number of blocks per layer were used. The grid size of $8$ points per function was used. The configurations were selected such that the distinct total number of parameters is obtained (small/medium/large models); the number of parameters is reported by the software (and is surprisingly different to the direct evaluation of the ``number of underlying functions'' times the ``grid size''). Each configuration was trained for $100$ passes through the training dataset, and the accuracy was printed out. The results are summarised in table \ref{FastKAN-CPU}. It can be seen that for any selected configuration, FastKAN struggles to achieve the training times of neural networks. It is interesting to note that the same configurations executed on the GPU gave the training times of $163.62$, $157.62$, and $112.34$ for the corresponding columns of table \ref{FastKAN-CPU}. 

\begin{table}
\begin{center}
\caption{Performance of FastKAN in \emph{Det4} example.}
\label{FastKAN-CPU}
\begin{tabular}{| l | l | r | r | r | } 
\hline
\multicolumn{2}{|l|}{config.} & $[256,8,16,1]$ & $[128,16,16,1]$ & $[32,16,16,1]$ \\ 
\multicolumn{2}{|l|}{num.\ par.} & $57497$ & $39857$ & $11921$ \\ 
\hline
\multicolumn{2}{|c|}{} & \multicolumn{3}{|c|}{accuracy} \\
\hline
\multirow{7}{*}{\rotatebox[origin=c]{90}{pass}} & $1$ & $0.794$ & $0.725$ & $0.422$ \\
& $2$ & $0.856$ & $0.827$ & $0.673$ \\
& $5$ & $0.922$ & $0.919$ & $0.810$ \\
& $10$ & $0.949$ & $0.946$ & $0.873$ \\
& $20$ & $0.957$ & $0.954$ & $0.893$ \\
& $50$ & $0.958$ & $0.955$ & $0.895$ \\
& $100$ & $0.958$ & $0.955$ & $0.895$ \\
\hline
\multicolumn{2}{|l|}{time (s)} & $358.25$ & $257.27$ & $213.42$ \\
\hline
\end{tabular}
\end{center}
\end{table} 

\subsubsection{Computational example Tetra}

Given that the previous subsection is entirely based only on one computational example, it is necessary to provide at least one more comparison. This subsection presents the results for the \emph{Tetra} example, also adding the comparison to PyTorch (version 2.5.1). The same hardware and software were used as above. 

The results are summarised in table \ref{tetrahedron}, where various configurations of neural networks and of KANs are shown. As before, the `config.'\ cell shows the configuration of the networks with the sizes of the hidden layers. All models are trained to the same accuracy of $98\%$, after achieving which the training is stopped. The time is shown in the table. It can be seen that PyTorch performs about $1.4$ times better than Keras in this example; more importantly, it can be seen that neural networks perform well in this example, and the performance gap between neural networks (Keras and PyTorch) and the authors' KAN implementation is not so drastic --- PyTorch is only about $4.6$ times slower when trained to the same accuracy on CPU. FastKAN on the other hand is still significantly slower than neural networks (the same grid size of $8$ points was used). The Python scripts that were used for the calculations can be downloaded from the authors' website\footnote{http://openkan.org/keras-fastkan-pytorch.html}.

In the authors' KAN implementation, three-layer KAN was trained: the first layer consisting of $40$ blocks with $2$ points per function, the second layer consisting of $12$ blocks with $16$ points per function, the third layer consisting of $4$ blocks with $22$ points per function. The number of threads was taken to be $10$ and the batch size was taken to be $7\mathrm{K}$.

\begin{table}
\begin{center}
\caption{Performance of neural networks (Keras and PyTorch), the authors' KAN implementation, and  FastKAN in \emph{Tetra} example.}
\label{tetrahedron}
\begin{tabular}{| l | l | r | c |}
\hline
software & config. & num. par. & time (s) \\
\hline
authors' KAN & see text & $9696$ & $1.78 \pm 0.07$ \\
\hline
\multirow{3}{*}{Keras} & $[64, 32, 32]$ & $4100$ & $15.66 \pm 0.23$ \\
& $[128, 64, 32]$ & $12132$ & $11.15 \pm 0.18$ \\
& $[128, 128, 64]$ & $26692$ & $13.57 \pm 0.46$ \\
\hline
\multirow{4}{*}{PyTorch} & $[128, 64, 32]$, GPU & $12132$ & $7.53 \pm 1.31$ \\
& $[128, 64, 32]$, CPU & $12132$ & $9.01 \pm 0.81$ \\
& $[128, 128, 32]$, CPU & $22436$ & $8.13 \pm 0.86$ \\
& $[256, 128, 32]$, CPU & $40484$ & $8.14 \pm 1.12$ \\
\hline
\multirow{4}{*}{FastKAN} & $[32, 32, 16, 4]$, GPU & $18156$ & $41.78 \pm 3.50$ \\
& $[32, 32, 16, 4]$, CPU & $18156$ & $55.03 \pm 14.55$ \\
& $[128, 64, 32, 4]$, CPU & $107868$ & $19.14 \pm 0.79$ \\
& $[32, 16, 8, 4]$, CPU & $9732$ & $86.48 \pm 32.44$ \\
\hline
\end{tabular}
\end{center}
\end{table}

\subsection{Testing of parallel implementation on FPGA board}

The proposed implementation was evaluated on Digilent Nexys A7-100T development board, which incorporates Xilinx Artix 7 FPGA. Owing to its affordability, this board is predominantly used for educational and prototyping purposes. It has the system clock with the operational frequency of $100\,\mathrm{MHz}$. The FPGA features $240$ DSP slices.

The considered example was \emph{Det3}. The dataset was generated dynamically during the training, ensuring that each successive record can be considered previously unseen. The computation of the error was integrated into the training procedure, with the most recent $256$ differences stored in a circular buffer and used to evaluate the prediction accuracy. Two-layer KANs were constructed, with $6$ blocks in the first layer and $3$ points per function, $1$ block in the second layer and $21$ points per function. After one pass through the training dataset, the model achieved the prediction accuracy exceeding $98\%$ on previously unseen data.

Both the register-transfer level (RTL) implementation and its corresponding C++ version are available online (see link above). As both implementations use integer arithmetic, their results are identical; the C++ version is particularly suitable for rapid experimentation with alternative network configurations.

Larger network configurations can be constructed; however, the key factor limiting parallel computation is the number of DSP slices. Modern high-performance boards designed for massive parallelism may possess e.g.\ $12{,}288$ DSP slices, which allows deployment of much larger models.

\section{Conclusions}
\label{sec:conclusion}

The present paper proposes and demonstrates four new concurrency-driven strategies to improve training of KANs. Three of these strategies are applicable to KANs with arbitrary basis functions, while all-integer division-free KANs for FPGAs rely on piecewise-linear basis functions. The weak and the strong scaling tests have been performed for the authors' implementation for CPUs and have shown that almost linear scaling (in terms of processing of the total computational work) is achieved when up to $16$ threads are used. However, since the ``training on disjoints'' strategy relies on merging of the models, some accuracy can be lost or gained due to the averaging. The accuracy loss can be mitigated by decrease of the batch size and the increase of the total computational work. The accuracy gain was observed when training to high accuracies (above $99\%$) was performed, where less total computational work was required to achieve the same accuracy in the multi-threaded case, leading to the speedup of $4$ on $4$ threads. Overall, in this case, it is useful to introduce the effective speedup measure --- to train the models on different number of threads to the same accuracy, irrespective of the total computational work, and then compare the speedup. 

The scaling tests led to two conclusions: (a) the number of threads has an optimal value if training is done to the same accuracy, i.e.\ small number of threads provides insufficient speedup while the large number of threads provides insufficient accuracy, (b) the decrease of the batch size leads to the improvement of accuracy; however, it simultaneously leads to larger number of model synchronisations that cost time. In practice, the number of threads should be between $4$ and $16$ for the optimal trade-off between accuracy and speedup; the batch size can be as large as $1\mathrm{M}$ for pre-trained models and should be below $5\mathrm{K}$ for non-pre-trained models, and a good strategy is probably changing (increasing) the batch size during the training.

The multi-threaded authors' C++ KAN implementation has been compared to FastKAN and to neural networks implemented in MATLAB, Keras, and PyTorch. The authors put significant effort into finding the best-performing neural network configurations in MATLAB by considering training to different accuracies (coarse/medium/fine models) and varying the number of layers, the number of neurons per layer, the activation function type, the number of training iterations, and even the floating-point precision (single/double). Nevertheless, the authors' KAN implementation provides considerably faster training time than neural networks, when trained on CPUs to the same accuracy. The observed performance gain was at least several times difference, depending on example, and exceeding $40$ times in some cases. Even in purely sequential form (one thread on a CPU), the implementation remains competitive with the GPU-accelerated training of neural networks, executed on a typical CUDA-enabled laptop. Of course, such comparison is not just ``MLP vs.\ KAN'' but heavily involves the training method of the networks (which are different for KANs and MLPs) and the implementation structure. The predictive performance across MATLAB, Keras, PyTorch, and FastKAN was consistent, with FastKAN demonstrating somewhat longer training times than neural networks.

These measurements, however, represent only a fraction of the attainable performance. The computational structure of the considered KAN training procedure is inherently aligned with hardware-level parallelism of FPGA and ASIC environments, where concurrency is not an optimisation but a structural feature. In the provided demo implementation for FPGAs, a single training record is processed within $14$ clock cycles, with additional $2$ clock cycles for data generation and $1$ clock cycle for resetting the state. At the operational frequency of $100\,\mathrm{MHz}$, this corresponds to a throughput exceeding $5.5\mathrm{M}$ training records per second. Notably, both the latency and the processing rate remain invariant with respect to the model size, provided that the hardware supports sufficient concurrency. This means that the large-scale example of this paper (\emph{Det5}), which was trained on an HPC cluster for $406\,\mathrm{s}$ (processing of $20\mathrm{M}$ training records), can be theoretically completed in less than $3.5\,\mathrm{s}$. This is not even accounting for the operational frequency of $100\,\mathrm{MHz}$ being significantly lower than that of the latest high-end FPGA boards.

Certain practical barriers remain for widespread industrial adoption of FPGA-based training for large KAN models. Large-scale designs require advanced hardware platforms and lengthy bitstream generation cycles, sometimes comparable in duration to compilation of complex CPU software. Moreover, reliable RTL development demands specialised expertise. These challenges, however, are organisational rather than theoretical. As industrial hardware capabilities continue to expand, the proposed training approach offers a scalable path towards increasingly efficient implementations.

Beyond raw training speed, considerations of maintainability and deployability are equally important. The presented C++ implementations are compact (below $550$ lines), free from third-party dependencies, and avoid language-specific constructs, facilitating portability and integration into diverse software ecosystems --- porting of the code to SystemVerilog and deployment on FPGAs is just one example. Meanwhile, users of alternative KAN implementations remain constrained by software-level design decisions, including dependency on specific runtime environments, which raise questions regarding deployability and integration of the models into non-Python-based applications.

The most comprehensive user-friendly multi-threaded version of the implementation is implemented in object-oriented C++ code by the authors\footnote{https://bitbucket.org/kolmogorov-arnold/kankan-11/src/master/}. In addition, the authors also provide an easier-to-read sequential C++ code without the use of classes\footnote{https://bitbucket.org/kolmogorov-arnold/kankan-9/src/master/}. The codes contain five built-in demonstrative examples. The governing idea behind the construction of the examples is easily-explainable intuitive and/or geometric meaning (rather than just large formulas), as well as mathematical complexity. Therefore, the examples are the prediction of the determinants (\emph{Det4}), the areas of random triangles, the values of random quadratic forms, the lengths of medians of random triangles (vector output), and the areas of faces of random tetrahedra (also vector output). For demo purposes, several examples have three and four KAN layers, designed to be convenient customisation templates. The authors also actively maintain the MATLAB implementation\footnote{https://github.com/mpoluektov/kan-polar} of KANs, which includes both the Gauss-Newton-based and the Newton-Kaczmarz-based training methods (as well as the accelerated Newton-Kaczmarz method with parallel evaluation of the basis functions), cubic spline and piecewise-linear basis functions, pre-training, built-in two-layer and three-layer architectures, and the data-driven solution of partial differential equations. Engineering-style documentation with many other benchmarks can be found on the authors' website\footnote{http://openkan.org}.

\section*{Acknowledgements}

MP acknowledges funding from Research England's ``Expanding Excellence in England'' (E3) fund via the ``Multi-scale Multi-disciplinary Modelling for Impact'' programme (M$^3$4Impact).

\appendix
\titleformat{\section}[hang]{\Large\bfseries\raggedright\sffamily}{Appendix \thesection}{1em}{}

\section{Rescaling of the model's parameters and optimal training}
\label{sec:rescaling}

In the general case of KANs with arbitrary basis, functions $g_{i,j}$ from equation \eqref{eq:layer} are represented via the product of basis functions $\psi_k$ and model parameters $G_{i,j,k}$:
\begin{equation}
g_{i,j} \left( y_j \right) = \sum_{k=1}^p G_{i,j,k} \psi_k \left( y_j \right) ,
\label{eq:funcBasis}
\end{equation}
where $p$ is the number of the basis functions per one underlying function of the block. The basis functions together cover domain $y \in \left[y_\mathrm{min}, y_\mathrm{max} \right]$, where values $y_\mathrm{min}$ and $y_\mathrm{max}$ are assumed for all layers except the first (for which the min/max of the inputs are directly given by the dataset). 

One aspect that has not been highlighted in the previous papers is that for each layer, or more precisely for each underlying function of a block of a layer, values $y_\mathrm{min}$ and $y_\mathrm{max}$ are also formally the parameters of the model. Previously, it has been suggested to take them equal to the min/max of the outputs of the modelled data. However, as will be shown below, this may not be the optimal choice.

Even though values $y_\mathrm{min}$ and $y_\mathrm{max}$ can be understood as some model parameters, they are not independent parameters --- one can change them arbitrarily without affecting the input-output mapping of the model. To prove this, a more general model for one block (relation \eqref{eq:layer} with decomposition \eqref{eq:funcBasis}) is considered with a constant parameter added to the sum of the functions. It is sufficient to consider the classical vector-input-scalar-output two-layer KAN. Such model is written as
\begin{equation}
z = G_0 + \sum_{j,k} G_{j,k} \psi_k \left( y_j \right) , \qquad
y_j = H_{j,0} + \sum_{l,r} H_{j,l,r} \phi_r \left( x_l \right) ,
\label{eq:twoLay}
\end{equation}
where $\vectorn{x}$ is the input vector, $\vectorn{y}$ is the intermediate (hidden) vector, $z$ is the output scalar (compare equations \eqref{eq:addends} and \eqref{eq:twoLay}; in the latter, functions $g_j$ and $h_{j,l}$ are represented via the basis functions and parameters $G_{j,k}$ and $H_{j,l,r}$, respectively). Basis functions $\psi_k$ and $\phi_r$ correspond to the outer and the inner layers, respectively, and are defined on domains $\left[y_\mathrm{min}, y_\mathrm{max} \right]$ and $\left[x_\mathrm{min}, x_\mathrm{max} \right]$, respectively. Obviously, $x_\mathrm{min}$ and $x_\mathrm{max}$ are directly given by the data. However, $y_\mathrm{min}$ and $y_\mathrm{max}$ can be freely chosen. 

It should be emphasised that care must be taken when comparing formulas to the previous publications, in particular \cite{Poluektov2023}, where the notation is slightly different (with $\tilde{y}$ instead of $z$, $\vectorn{\theta}$ instead of $\vectorn{y}$, and $y$ instead of $z^*$); the change of notation was introduced for more natural alignment with the explanations of the present paper.

Suppose that a linear mapping is applied to the intermediate vector: $\vectorn{y}' = \alpha \vectorn{y} + \beta$. The elements of the new intermediate vector have limits $y_\mathrm{min}' = \alpha y_\mathrm{min} + \beta$ and $y_\mathrm{max}' = \alpha y_\mathrm{max} + \beta$. Substituting the mapping into model \eqref{eq:twoLay} results in
\begin{equation}
z = G_0 + \sum_{j,k} G_{j,k} \psi_k \left( \alpha^{-1} y_j' - \alpha^{-1}\beta \right) , \qquad
y_j' = \alpha H_{j,0} + \beta + \sum_{l,r} \alpha H_{j,l,r} \phi_r \left( x_l \right) .
\end{equation}
New basis functions and parameters can be defined as
\begin{equation}
\xi_k \left( y_j' \right) = \psi_k \left( \alpha^{-1} y_j' - \alpha^{-1}\beta \right) , \qquad
H_{j,0}' = \alpha H_{j,0} + \beta, \qquad
H_{j,l,r}' = \alpha H_{j,l,r} ,
\end{equation}
resulting in 
\begin{equation}
z = G_0 + \sum_{j,k} G_{j,k} \xi_k \left( y_j' \right) , \qquad
y_j' = H_{j,0}' + \sum_{l,r} H_{j,l,r}' \phi_r \left( x_l \right) .
\label{eq:twoLayScal}
\end{equation}
Resulting model \eqref{eq:twoLayScal} has exactly the same structure as original model \eqref{eq:twoLay}, but with the second-layer basis functions defined on rescaled domain $\left[y_\mathrm{min}', y_\mathrm{max}' \right]$. As evident from the equations, such rescaling leads only to the change of the parameters of the inner layer. Since $\alpha$ and $\beta$ are arbitrary, one can perform arbitrary rescaling of the intermediate vector, exactly preserving input-output relationship $z\left(\vectorn{x}\right)$ of the model. Thus, $y_\mathrm{min}$ and $y_\mathrm{max}$ are not independent model parameters, and only the multipliers in front of the basis functions constitute the model parameters (moreover, they are not totally independent either, as shown for one block, see \cite{Poluektov2020}).

For the two-layer model given by system \eqref{eq:twoLay}, the parameters' updates are written as \cite{Poluektov2023}
\begin{align}
  &H_{j,l,r}^\mathrm{new} = H_{j,l,r}^\mathrm{old} + \mu\zeta^{-1} \left(z^* - z\right)\frac{\partial z}{\partial H_{j,l,r}} , \\
  &G_{j,k}^\mathrm{new} = G_{j,k}^\mathrm{old} + \mu\zeta^{-1} \left(z^* - z\right)\frac{\partial z}{\partial G_{j,k}} , \\
  &\zeta = \sum_{j,l,r} \left(\frac{\partial z}{\partial H_{j,l,r}}\right)^2 + \sum_{j,k} \left(\frac{\partial z}{\partial G_{j,k}}\right)^2 .
\end{align}
where $\mu \in \left(0,1\right]$ is the numerical damping parameter. Variable $\zeta$ is the squared norm of the gradient of $z$ with respect to the parameters. Suppose that a linear mapping of the intermediate vector has been performed. Substituting the relation between $H_{j,l,r}'$ and $H_{j,l,r}$ results in
\begin{align}
  &H_{j,l,r}^{\prime\;\mathrm{new}} = H_{j,l,r}^{\prime\;\mathrm{old}} + \mu \alpha^2 \zeta^{-1}  \left(z^* - z\right) \frac{\partial z}{\partial H_{j,l,r}'} , \label{eq:iter0b} \\
  &G_{j,k}^\mathrm{new} = G_{j,k}^\mathrm{old} + \mu\zeta^{-1} \left(z^* - z\right) \frac{\partial z}{\partial G_{j,k}} ,  \\
  &\zeta = \alpha^2 \sum_{j,l,r} \bigg(\frac{\partial z}{\partial H_{j,l,r}'}\bigg)^2 + \sum_{j,k} \bigg(\frac{\partial z}{\partial G_{j,k}}\bigg)^2 . \label{eq:iter0e}
\end{align}
This is contrasted with the direct application of the NK step to the two-layer model given by system \eqref{eq:twoLayScal}, which consists in updates
\begin{align}
  &H_{j,l,r}^{\prime\;\mathrm{new}} = H_{j,l,r}^{\prime\;\mathrm{old}} + \nu \zeta^{\prime \, -1}  \left(z^* - z\right) \frac{\partial z}{\partial H_{j,l,r}'} , \label{eq:iter1b} \\
  &G_{j,k}^\mathrm{new} = G_{j,k}^\mathrm{old} + \nu\zeta^{\prime \, -1} \left(z^* - z\right)\frac{\partial z}{\partial G_{j,k}} , \\
  &\zeta' = \sum_{j,l,r} \bigg(\frac{\partial z}{\partial H_{j,l,r}'}\bigg)^2 + \sum_{j,k} \left(\frac{\partial z}{\partial G_{j,k}}\right)^2 , \label{eq:iter1e} 
\end{align}
where $\nu \in \left(0,1\right]$ is another numerical damping parameter. Update schemes \eqref{eq:iter0b}-\eqref{eq:iter0e} and \eqref{eq:iter1b}-\eqref{eq:iter1e} are clearly different even if one tries to match them by choosing specific numerical damping parameters. In particular, if it is required that the update of the outer layer parameters is the same in both schemes, then $\mu\zeta^{-1} = \nu\zeta^{\prime \, -1}$ must be enforced by the appropriate choice of $\mu$ and $\nu$. However, the update of the inner layer parameters will then have an extra multiplier $\alpha^2$ in the first scheme compared the second scheme.

Because models \eqref{eq:twoLay} and \eqref{eq:twoLayScal} have the same input-output relationship $z\left(\vectorn{x}\right)$, there is obvious freedom in applying the NK update scheme to either model \eqref{eq:twoLay} or model \eqref{eq:twoLayScal}. This however results in either presence or absence of multiplier $\alpha^2$ in the update of the inner layer parameters. Furthermore, factor $\alpha$ is arbitrary. This means three things. First, the numerical damping parameters for the layers are different and independent, giving the general update scheme with two numerical dampings, $\mu$ and $\nu$, also splitting variable $\zeta$ into two variables:
\begin{align}
  &H_{j,l,r}^\mathrm{new} = H_{j,l,r}^\mathrm{old} + \nu\zeta_H^{-1} \left(z^* - z\right)\frac{\partial z}{\partial H_{j,l,r}} , \label{eq:iter2b} \\
  &G_{j,k}^\mathrm{new} = G_{j,k}^\mathrm{old} + \mu\zeta_G^{-1} \left(z^* - z\right)\frac{\partial z}{\partial G_{j,k}} , \\
  &\zeta_H = \sum_{j,l,r} \left(\frac{\partial z}{\partial H_{j,l,r}}\right)^2 , \qquad
  \zeta_G = \sum_{j,k} \left(\frac{\partial z}{\partial G_{j,k}}\right)^2 . \label{eq:iter2e} 
\end{align}
Second, rescaling of the intermediate vector and applying the NK update scheme to the rescaled model is equivalent to changing the numerical damping parameters independently in the original model. Third, there must exist optimal ratio between the numerical damping parameters (or equivalently optimal range $\left[y_\mathrm{min}', y_\mathrm{max}' \right]$ of the intermediate variables). The reason for this is evident from scheme \eqref{eq:iter0b}-\eqref{eq:iter0e} --- very large $\alpha$ leads to very slow (small) updates of the outer layer parameters, while very small $\alpha$ leads to very slow (small) updates of the inner layer parameters --- both are undesired behaviours as the intention of having the two-layer model with initially-random parameters is the synergy between the updates of both layers.

Finally, for practical purposes, it is useful to quantify the values of variable $\zeta$ of an individual layer in scheme \eqref{eq:iter2b}-\eqref{eq:iter2e} when applied to piecewise-linear basis functions. In this case, each underlying function of each block of a layer has only two parameters that are updated. The derivatives with respect to these parameters are $\left(1-f\right)$ and $f$ (see equations \eqref{R}-\eqref{eq:g-interp}). Obviously, since $f \in [0,1)$, then $1/2 \leq \left(1-f\right)^2 + f^2 \leq 1$. Since the number of the underlying functions of a layer is $mn$ (see equation \eqref{eq:layer}), then $mn/2 \leq \zeta \leq mn$. Thus, for a shorter computational code, variable $\zeta$ can be approximated as $mn$, although leading to a slightly less efficient update.

\bibliographystyle{unsrt}
\bibliography{refs}

\end{document}